\pdfoutput=1

\documentclass[11pt]{article}

\usepackage{EMNLP2023}

\usepackage{times}
\usepackage{latexsym}
\usepackage{longtable}

\usepackage[T1]{fontenc}

\usepackage[utf8]{inputenc}

\usepackage{microtype}

\usepackage{inconsolata}
\usepackage{graphicx}
\usepackage{multirow}
\usepackage{amssymb}
\usepackage{caption}
\usepackage{subcaption}
\usepackage{booktabs}
\usepackage{algpseudocode}
\usepackage{algorithm}
\usepackage{tabularx}
\usepackage{xcolor,colortbl}

\title{Mastering the Task of Open Information Extraction with Large Language Models and Consistent Reasoning Environment}

\author{Ji Qi$^1$\footnotemark[1], Kaixuan Ji$^1$\footnotemark[1], Xiaozhi Wang$^1$, Jifan Yu$^1$, Kaisheng Zeng$^{1}$\\ \textbf{Lei Hou}$^{1}$, \textbf{Juanzi Li}$^{1}$, \textbf{Bin Xu}$^{1}$\footnotemark[2] \\
  $^1$Department of Computer Science and Technology, Tsinghua University \\
  \texttt{\{qj20,jkx19\}@mails.tsinghua.edu.cn}
}

\begin{document}
\maketitle
\renewcommand{\thefootnote}{\fnsymbol{footnote}}
\footnotetext[1]{Equal Contribution.}
\footnotetext[2]{Corresponding author.}
\renewcommand*{\thefootnote}{\arabic{footnote}}

\begin{abstract}
Open Information Extraction (OIE) aims to extract objective structured knowledge from natural texts, which has attracted growing attention to build dedicated models with human experience.
As the large language models (LLMs) have exhibited remarkable in-context learning capabilities, a question arises as to whether the task of OIE can be effectively tackled with this paradigm?
In this paper, we explore solving the OIE problem by constructing an appropriate reasoning environment for LLMs.
Specifically, we first propose a method to effectively estimate the discrepancy of syntactic distribution between a LLM and test samples, which can serve as correlation evidence for preparing positive demonstrations.
Upon the evidence, we introduce a simple yet effective mechanism to establish the reasoning environment for LLMs on specific tasks.
Without bells and whistles, experimental results on the standard CaRB benchmark demonstrate that our $6$-shot approach outperforms state-of-the-art supervised method, achieving an $55.3$ $F_1$ score.
Further experiments on TACRED and ACE05 show that our method can naturally generalize to other information extraction tasks, resulting in improvements of $5.7$ and $6.8$ $F_1$ scores, respectively.
\end{abstract}

\section{Introduction}

In contrast to the information extraction tasks (\emph{e.g.}, RE~\cite{li2022graph} and EE~\cite{hao2023devil}) that acquire knowledge aligned with subjective ontologies, Open Information Extraction (OIE) focuses on extracting all potential objective knowledge of N-tuples from natural texts, which benefits various domains and applications~\cite{gashteovski2020aligning,pei2022use}.
Starting from heuristic approaches learned with linguistic features on noisy web corpora~\cite{etzioni2008open,angeli2015leveraging}, the paradigm of OIE models has shifted to design neural models that are trained from scratch on parallel supervision~\cite{stanovsky2018supervised,cui2018neural}. Recently, the state-of-the-art methods have showcased the advantages of fine-tuning models based on pre-trained backbones and labeled datasets~\cite{kolluru2020openie6,qi2022syntactically}.
The community is devoted to training dedicated models on high-quality supervision.

\begin{figure}[pt]
    \centering
    \hspace{-0.2cm}\includegraphics[scale=0.5]{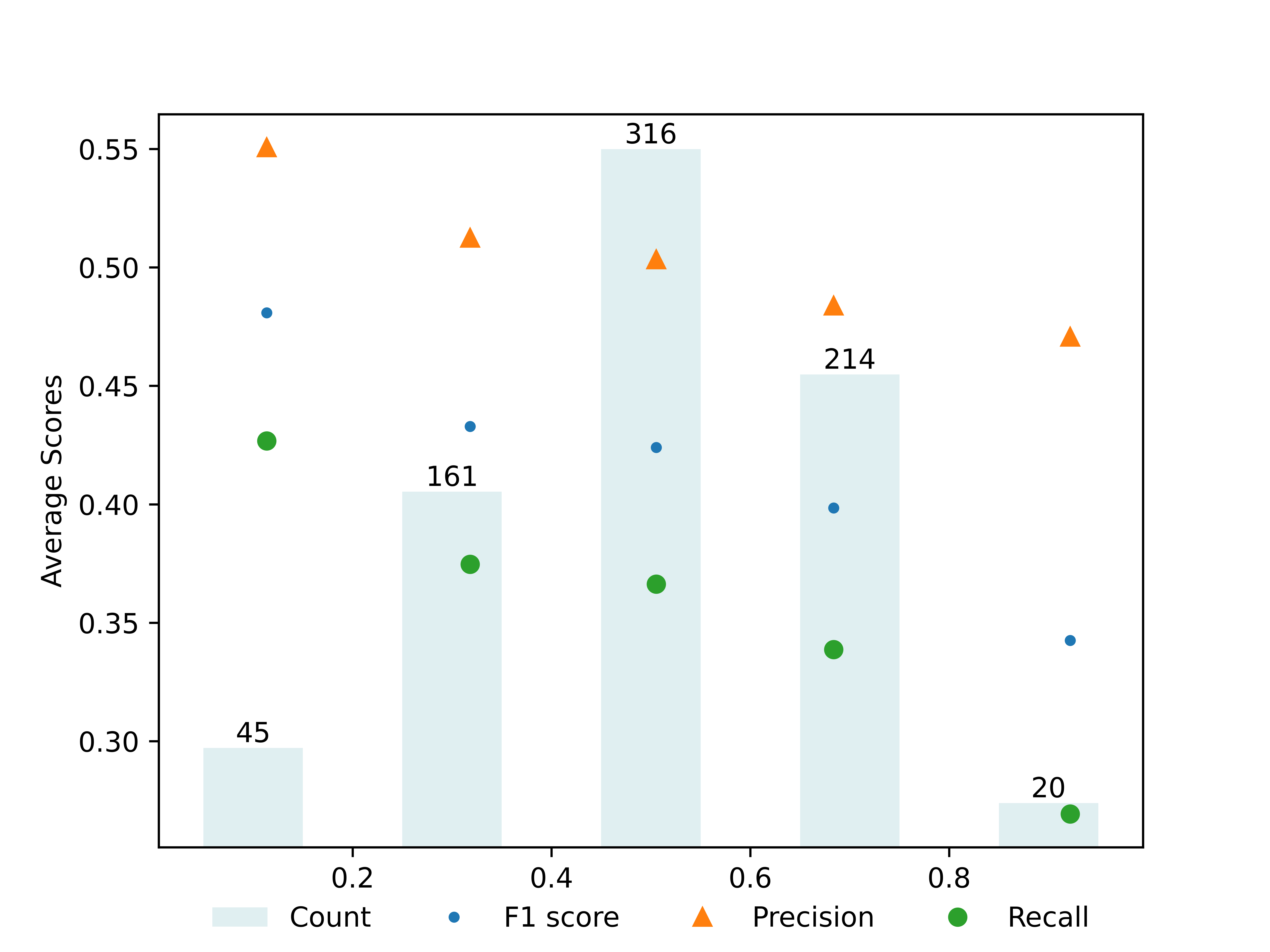}
    \caption{The OIE performance of ChatGPT decreases as the syntactic discrepancy between the test samples and ChatGPT increases.}
    \label{fig:example-oie}
\end{figure}

On the other hand, Large Language Models (LLMs)~\cite{brown2020language,ouyang2022training,chowdhery2022palm,chung2022scaling,touvron2023llama} have shown the remarkable in-context learning abilities that generate responses by only providing proper prompts with a few demonstrations without any tuning of parameters. Built upon~\cite{ouyang2022training}, ChatGPT is one of the most representative LLMs that train a decoder-only Transformer~\cite{vaswani2017attention} using reinforcement learning from human feedback (RLHF)~\cite{christiano2017deep}. As an interactive assistant, ChatGPT is shifting the solution of downstream NLP tasks into the prompts-driven paradigm~\cite{zhang2022would,qin2023chatgpt,mitrovic2023chatgpt,guo2023close}.

As an objective factual task highly correlated with sentence structures and semantics, a natural idea is whether LLMs can be employed as the efficient few-shot OIE extractors? To answer this question, a crucial challenge lies in constructing an appropriate demonstration environment for LLMs to perform reasoning, as in most cases we have limited knowledge about the task-specific information that LLMs may have missed during training.

As shown in Figure~\ref{fig:example-oie}, we propose an estimation approach to measure the discrepancy of syntactic distributions (introduced in Sec~\ref{subsec:estimation_method}) between test samples and ChatGPT, and further explore the correlation between this discrepancy and the OIE performance of ChatGPT.
We randomly sample 256 sentences with extractions from the ROBUST~\cite{qi2023preserving}, an OIE benchmark annotated by human experts. For each sentence, we compute the syntactic discrepancy (\textit{the horizontal axis}) between the sentence and ChatGPT, as well as the OIE performance (\textit{the vertical axis}) by prompting ChatGPT on the sentence\footnote{We divide all samples into 5 intervals according to the discrepancies and calculate the means to avoid abnormal values.}.
The observation demonstrates a strong correlation between the syntactic discrepancy and extraction performance. It suggests that we may significantly improve the performance by mitigating the discrepancy and establishing a consistent reasoning environment on this task.

In this paper, we explore the approach of constructing a consistent reasoning environment by mitigating the distributional discrepancy between test samples and LLMs, thereby improving the few-shot reasoning capability of LLMs on specific tasks.
Specifically, we first propose a method to estimate the discrepancy of syntactic distributions between the black-box LLM and test samples by employing a discrepancy metric.
The correlation between this discrepancy and the performance of LLM on specific tasks can be naturally validated following the estimation.
Based on the validation, we further introduce a simple yet effective mechanism to establish a consistent reasoning environment for LLMs on a specific task.
The environment comprises a majority of examples that are similar to the test samples in distribution and a few variants, ensuring both consistency and diversity.
By replacing the discrepancy metric, the approach can be naturally transferred to other NLP tasks.
The mechanism can thus serve as a guideline for preparing the reasoning environment, whether it is selected from a candidate set or curated from scratch.

We conduct extensive experiments on the standard OIE benchmark CaRB, including few-shot experiments ranging from 3 to 7 and control variable experiments, manipulating the size of candidate set samplings.
The experimental results show that our proposed method effectively improves the few-shot reasoning performance of LLMs. Our 6-shot result surpasses the current state-of-the-art supervised model, achieving a $55.3$ $F_1$ score.
In addition, we further conduct experiments on two ontology-specified information extraction tasks, Relation Extraction (RE) and Event Extraction (EE), by replacing the discrepancy metric to a content-based measurement.  The results show that our method gains improvements of 5.7 Micro $F_1$ scores and 6.8 $F_1$ scores, respectively.

\section{Methodology}

\begin{figure*}[h]
    \centering
    \includegraphics[scale=0.85]{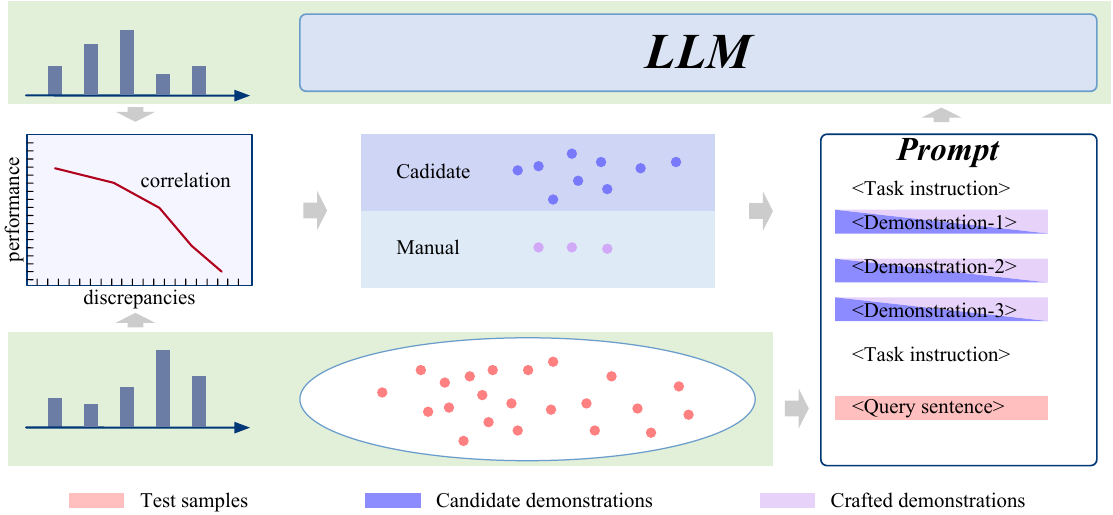}
    \caption{Overview of the proposed framework. Given a specific task (\emph{e.g.,} OIE), we first calculate the distributional discrepancies (\emph{e.g.,} syntactic distribution) between test samples and a black-box LLM, then further derive the correlation between the model performance and the discrepancy. Guided by this correlation, the reasoning environment is constructed with a majority of positive demonstrations selected from candidates or crafted manually.}
    \label{fig:framework}
\end{figure*}

The overall framework is shown in Figure~\ref{fig:framework}.
We first introduce the detailed approach for estimating the distributional discrepancy between a black-box LLM and test samples on a specific task, then derive the correlation between the model performance and the discrepancy. Upon this correlation, we further present the mechanism of constructing a reasoning environment efficiently with a majority of positive demonstrations selected from candidates or crafted manually.

\subsection{Estimating the Discrepancy of Syntactic Distribution (D-SD) to LLMs}
\label{subsec:estimation_method}

Given the black-box LLMs that do not provide any training details, it is essential to estimate the distributional discrepancy between the LLMs and test samples for investigating what they don't know about the specific task, and further enhance their few-shot reasoning capabilities by mitigating this discrepancy.
The core idea behind our approach is straightforward. Intuitively, given the semantics of a sentence, the sentences reconstructed by a LLM reflect the linguistic features learned by the model related to that semantics.
In this way, we can calculate the performance of the LLM on a specific task on the original sentences, as well as the discrepancies between the source sentences and the target sentences.
Given the source sentences in a test set, the core problem transforms into how to eatable the LLMs to reconstruct a set of target sentences based on the semantics of each source sentence, and compute an unbiased estimation of the discrepancies.

Inspired by \citet{qi2023preserving},
we develop the following approach to simulate the aforementioned estimation process.
Given a test set $S=\{(x_i,y_i)\}^k$ consisting of $k$ pairs of samples, each representing an input sentence and its golden output, we first manually create $m$ paraphrases for each sentence $x_i$. The test set, together with these manually crafted paraphrases forms the source cliques $\{C_i|C_i=(x^0_i,x^1_i, ..., x^m_i, y_i)\}$\footnote{We use $x^0_i$ to denote the original $x_i$ for convenience.}. Note that each clique represents the same semantic meaning of its sentences.
For each source clique $C_i$, we then prompt the LLM to paraphrase each sentence in it to produce the target clique $C'_i=(x'^0_i,x'^1_i, ..., x'^m_i)$.
The target clique fairly reflects the linguistic features of the LLM under the given semantics, with the semantics being consistent with the source clique.
For each sentence $x^j_i$ in the source clique $C_i$, we calculate its syntactic discrepancy to the LLM as:

\begin{equation}
    d(x^j_i) = \frac{1}{m}\sum_{l\neq j}f_d(x^j_i, x'^l_i),
\end{equation}
where we use the Hierarchically Weighted Syntactic (HWS) distance \citep{qi2023preserving} to implement the discrepancy metric $f_d$.
The results of $[d(x^0_1), ..,  d(x^j_i), .., d(x^m_k)]$ serve as the discrepancy of syntactic distributions between the test set and LLM.

Based on this estimation, we can naturally validate the correlation between the discrepancy and the LLM performance on a specific task, \emph{i.e.,} prompting the LLM to generate task-specific output $y'^0_i$ for each source sentence $x^0_i$, and calculate the prediction performance $p(x^0_i)$ against the corresponding golden output $y^0_i$.
We obtain this correlation by observing the trends or calculating the Pearson's Correlation Coefficient between the two variables based on the resulting $\{(d(x^0_i), p(x^0_i)\}$.
The designed prompts for generating paraphrases from LLMs are available in Appendix~\ref{app:prompt_paraphrase}.

\subsection{Preparing the Consistent Reasoning Environment for LLMs}
\label{subsec:preparing_mechanism}

Guided by the estimation of distributional discrepancy and the validation of performance correlation, we can naturally prepare the positive reasoning environment by constructing demonstrations that minimize the discrepancy with each test sample.
However, we find that arbitrarily providing positive demonstrations according to the discrepancy metric does not result in the optimal performance in a realistic reasoning scenario (as indicated by the highest performance in Figure~\ref{fig:example-oie}, which falls below the supervised SOTA).
Therefore, we propose a construction mechanism that leverages positive demonstrations to dominate the environment while simultaneously introducing variants, ensuring both consistency and diversity.

In the scenario of given $n$ samples of the candidate set $\{(x^s_i, y^s_i)\}$, for a query sentence $q$, we first calculate the discrepancy value $f_d(q, x^s_i)$ between the query and each candidate sentence based on the discrepancy metric. Subsequently, we sample the few-shot demonstrations using the reciprocals of these normalized discrepancies as a probability distribution. These results ensure both a minimum discrepancies as the mean and a varying variance.
In the scenario of manually crafting a reasoning environment, for a given query sentence $q$, we refer to the validated discrepancy metric and create the majority of demonstrations in a way that minimizes this measure, and add a few demonstrations with large discrepancy to complete the construction of the environment.
We empirically set the ratio to $8:2$ for the two parts of the demonstrations.
The prompts designed for the OIE task and partial results of selecting examples based on syntactic discrepancy metric are available in Appendix~\ref{app:prompt_oie} and Appendix~\ref{app:case}, respectively.

\section{Experiments}

To evaluate the effectiveness of the proposed approach, we conduct extensive experiments on the standard OIE benchmark, CaRB~\cite{bhardwaj2019carb}. Our experiments include the estimation of distributional discrepancy with corresponding validation of performance correlation, and the few-shot evaluations based on the proposed mechanism.

Due to the applicability of the proposed method, we further conduct experiments by replacing the discrepancy metric on two ontology-specified information extraction tasks: RE and EE.

\subsection{Experiments on OIE Task}

\subsubsection{Experimental Settings}

We use CaRB, a standard OIE benchmark crowd sourced by annotators consisting of 1272 sentences with 5262 tuple extractions for our test set.\footnote{We follow \citet{qi2023preserving} and remove the 10 sentences with no extractions}. We follow the previous works and report precision, recall and $F_1$ score based on CaRB scorer.
We evaluate our method by sampling demonstrations from ROBUST~\cite{qi2023preserving}, a human-annotated OIE testbed consisting of $1272$ where each clique comprises a CaRB sentence with multiple synonymous paraphrases.
We perform multiple times of sampling from ROBUST to obtain the demonstrations candidate sets of different size $S$. For each candidate set, we conduct experiments with 3 to 7 shots. When sampling demonstrations for a query sentence, we exclude cliques from the candidate set that contained the query sentence.
Note that the larger size of demonstrations candidate set implies a closer approximation to humans manually crafting the demonstrations accordingly.



For the implementation of LLMs, we use the representative model ChatGPT with its official API of "turbo-3.5" provided by OpenAI\footnote{Our experimental period is 2023.03.07-2023.06.08}.
We use the Hierarchically Weighted Syntactic (HWS) distance~\cite{qi2022syntactically} as the metric.
We compare our few-shot approach with supervised models covering the mainstream structures, including the heuristic model \textit{OpenIE4}~\cite{mausam2016open}, the neural models \textit{SpanOIE}~\citep{zhan2020span} and \textit{RnnOIE}\citep{stanovsky2018supervised}, and the pre-training-based models \textit{OpenIE6}~\cite{kolluru2020openie6}.

\subsubsection{The Estimation of Syntactic Discrepancy}
\label{subsubsec:estimate_syntactic_discrepancy}

To validate the significance of D-SD, we estimate the syntactic discrepancy between the CaRB set and ChatGPT, and further derive the correlation between the OIE performance of ChatGPT and the discrepancies.
We first randomly sample $256$ source cliques from ROBUST, where each clique consists of syntactically diverse sentences conveying the same knowledge meaning. We then obtain the corresponding target cliques by prompting ChatGPT to paraphrase the source clique, and prompt ChatGPT to generate OIE extractions for all source cliques.
For each sentence in source cliques, we calculate the extracting performance on the sentence and the averaged syntactic distance between the sentence and its corresponding target cliques as introduced in Sec~\ref{subsec:estimation_method}. We divide all samples into 5 intervals according to the discrepancies and calculate the means of performance to avoid abnormal values.

The results including $3$ CaRB scores are shown in Figure~\ref{fig:example-oie}. We can see that the results demonstrate a strong correlation between the discrepancy of syntactic distribution (D-SD) and the OIE performance of ChatGPT. This observation effectively indicates that the LLM OIE performance can be significantly improved by mitigating this syntactic discrepancy.

\begin{table*}
 \setcounter{table}{1}
  \centering
  \begin{tabular}{l>{\centering}p{2.2cm}>{\centering}p{1.6cm}>{\centering}p{1.6cm}>{\centering}p{1.6cm}c}
    \toprule
      \begin{minipage}{1.6cm}\ \\\textbf{Model}\end{minipage} & \textbf{ChatGPT} & \multicolumn{4}{c}{\textbf{ChatGPT}} \\
    \cline{3-6}
        & \textbf{Random} & 50 & 200 & 1272 & 4932 \\
    \hline
      $Precision$ & 60.7 & 60.5 & 61.1 & 60.0 & \textbf{62.3} \\
      $Recall$    & 41.2 & 42.6 & 43.4 & 42.4 & \textbf{45.8} \\
      $F_1$        & 49.1 & 50.0 & 50.8 & 49.7 & \textbf{52.7} \\
    \hline
     \textit{Average Syntactic Distance} & - & 0.388 & 0.282 & 0.178 & 0.142 \\
    \hline
  \end{tabular}
  \caption{$3$-shot results of ChatGPT with continuously increasing the candidate size of demonstrations. We also calculate the average syntactic distance between the test sets and demonstrations to investigate the correlation.}
  \label{tab:result-openie-ablation}
\end{table*}

\begin{table}
 \setcounter{table}{0}
  \centering
  \begin{tabular}{lccc}
    \toprule
    \textbf{Model}   & \textbf{P}  & \textbf{R}  & $\mathbf{F_1}$   \\
    \midrule
    OpenIE4   & 61.0 & 48.3 & 53.9 \\
    SpanOIE   & 31.3 & 42.2 & 36.0 \\
    RnnOIE    & 49.3 & 49.5 & 49.4 \\
    OpenIE6   & 60.9 & 50.5 & 55.2 \\
    \midrule
    ChatGPT ($n$=3, $S$=4932) & 62.3 & 45.8 & 52.7 \\
    ChatGPT ($n$=4, $S$=4932) & 62.9 & 47.2 & 53.9 \\
    ChatGPT ($n$=5, $S$=4932) & 63.5 & 48.0 & 54.7 \\
    {ChatGPT ($n$=6, $S$=4932)} & \textbf{63.2} & 49.1 & \textbf{55.3} \\
    ChatGPT ($n$=7, $S$=4932) & 62.7 & 48.9 & 55.0 \\
    \bottomrule
  \end{tabular}
  \caption{$n$-shot results of ChatGPT compared to existing representative OIE models. Our $6$-shot result outperforms the fully supervised SOTA model OpenIE6.}
  \label{tab:result-openie}
\end{table}

\subsubsection{Experimental Results}

We first fix the size of candidate set $S$ to be 4932, which is the largest demonstration candidate corpus set and set $n$ to be 3 to 7.
To fairly compare with existing models, we perform experiment and compute average performance on the full set of CaRB.
The experimental results are shown in table~\ref{tab:result-openie}.
The results show that, by providing appropriate few-shot demonstrations, our method enables ChatGPT to outperform all the rule-based and non-pretrained-model-based methods by a considerable margin (1.4\% in F1 score).
With the $6$-shot setting, our method surpass the current state-of-the-art fully supervised model, achieving a $55.3$ F1 score.

We then varies the size of demonstrations candidate set to see the performance changes. We fix $n=3$ and continuously expand the size of demonstrations candidate set. The results are shown in table~\ref{tab:result-openie-ablation}.
Our method can obtain growing performance when linearly increases the size of candidate demonstrations set, and achieve the best $F_1$ score with the largest size of $4932$.
We also calculate the average syntactic distance between test sets and demonstrations using HWS distance. We find that as the demonstrations become more syntactically similar, ChatGPT tends to perform better performance. These results align with our intuition and indicate the practicality of our method.

\subsubsection{Case Study}

We also perform a case study to illustrate the issues encountered by ChatGPT during OIE prediction and demonstrate the effectiveness of our proposed method.
For the following sentence with its one corresponding knowledge tuple, we compare the ChatGPT results obtained by selecting 3 demonstrations from the candidate set of size 50 (almost the same as random selection) and from the candidate set of size 4392:


\begin{itemize}
    \item \textbf{Text}: \textit{About 60 \% of the work force will continue with Gillette or transfer to Twins Pharmaceuticals, the company said.}
    \item \textbf{Tuple-1}: (said, the company, About 60 \% of the work force will continue with Gillette or transfer to Twins Pharmaceuticals)
\end{itemize}

With nearly randomly picked demonstration (see Appendix~\ref{app:subsubsec:demos_oie_random}), ChatGPT fails to extract the corresponding tuple-1 from the given sentence.

On the other hand, experiments shows that our methods enables ChatGPT to extract this tuple correctly.
Our method provid more effective demonstrations (see Appendix~\ref{app:subsubsec:demos_oie_hws} for details) to the current query sentence to enhance to extraction.
For example, one of the selected demonstrations based on our method is:
\begin{itemize}
    \item \textbf{Text}: \textit{The fitness craze itself has gone soft , the survey found.}
    \item \textbf{Tuple-1}\footnote{Here we omit other irrelevant tuples}: (found, the survey, The fitness craze itself has gone soft)
\end{itemize}

\begin{table*}[pt]
 \setcounter{table}{2}
  \centering
  \begin{tabular}{lc>{\centering}p{1.2cm}>{\centering}p{1.2cm}>{\centering}p{1.2cm}>{\centering}p{1.2cm}>{\centering}p{1.2cm}c}
    \toprule
      \begin{minipage}{1.6cm}\ \\\textbf{Model}\end{minipage} & \textbf{KLG}~\citep{li2022reviewing} & \multicolumn{6}{c}{\textbf{ChatGPT}} \\
    \cline{3-8}
        & - & 100 & 500 & 1000 & 2000 & 5000 & 20000 \\
    \hline
        Micro $\mathbf{F_1}$ & 75.6 & 58.8 & 62.8 & 67.6 & 64.4 & 64.8 & 64.6 \\
    \hline
  \end{tabular}
  \caption{The 30-shot results of ChatGPT on the TACRED benchmark compared to the supervised SOTA KLG. We randomly sample 500 examples in the test set for evaluation. The demonstrations are selected based on our proposed mechanism in the provided training set.}
  \label{tab:result-re}
\end{table*}

Both our demonstration and the test sample have the syntactic structure of object clause, which are directly reflected in their golden extractions. Therefore, our demonstration depicts the extraction strategy under this scenario so that ChatGPT can extract the tuple correctly which is neglected by ChatGPT with random demonstration.

\subsection{Experiments on RE Task}

In contrast to OIE, the RE task aims to extract ontology-specified relations aligned with user's subjective intentions. Therefore, we evaluate our proposed framework by changing the discrepancy metric to the content-based measurement. In this paper, we use CaRB scorer, a simple metric that measure the correctness of extracted sub-contents for a given sentence as the implementation.

\subsubsection{Experimental Settings}

We use TACRED~\citep{zhang2017position}, a widely-used RE benchmark consisting of 68124, 22631 and 15509 samples for the splits of train, development and test for our evaluation. In each sample two entities are marked and the relation between them is to be categorized into 42 classes. Following the previous works, we report micro-F1 score.
We compare our few-shot approach with the current supervised SOTA model KLG~\citep{li2022reviewing}, which evaluates on the whole test set.

We use the same ChatGPT version\footnote{Our exprimental period is 2023.04.11-2023.06.08.} as the standard implementation of LLM.
For the content-based discrepancy measurement, we use IMOJIE~\citep{kolluru2020imojie} to extract tuples from sentences and calculate CaRB $F_1$ score to measure the content discrepancy between sentences.
The prompt for the RE task is available at Appendix~\ref{app:prompt_re}.

\subsubsection{The Estimation of Content Discrepancy}

\begin{figure}[pt]
    \centering
    \hspace{-0.2cm}\includegraphics[scale=0.5]{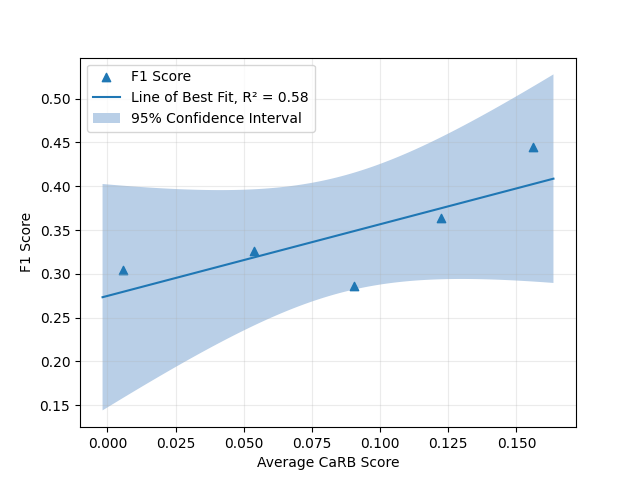}
    \caption{Validating the correlation between content-based distributional discrepancy and the relation extraction performance.}
    \label{fig:example-re}
\end{figure}

To validate the correlation between the selected discrepancy metric and the RE performance of ChatGPT, we implement the proposed estimation method introduced in Sec~\ref{subsec:estimation_method} by changing the discrepancy metric to the content-based measurement CaRB scorer.
Different from the idea that let LLMs to paraphrase given sentences, for each RE sample $(s_i, \langle h_i,r_i,t_i \rangle)$ indicating a sentence-extraction pair, we prompt ChatGPT to make multiple sentences $\{s'_i\}$ by giving the head-tail entities $h_i, t_i$. We then calculate the average $F_1$ CaRB score as the discrepancy value $d(s_i)$ on the tuples extracted based on IMOJIE from original sentence $s_i$ and target sentences $\{s'_i\}$. We further prompt ChatGPT to extract the relation $r'_i$ by giving $(s_i, \langle h_i, t_i\rangle)$ and calculate the performance of Micro F1 $p(s_i)$.

\begin{table*}[pt]
    \setcounter{table}{4}
    \begin{tabular}{lc>{\centering}p{1.2cm}>{\centering}p{1.2cm}>{\centering}p{1.2cm}>{\centering}p{1.2cm}c}
    \toprule
      \begin{minipage}{1.6cm}\ \\\textbf{Model}\end{minipage} & \textbf{SaliencyED}~\citep{liu2022saliency} & \multicolumn{5}{c}{\textbf{ChatGPT}} \\
    \cline{3-7}
        & - & 50 & 500 & 1500 & 5000 & 15000 \\
    \hline
        $Precision$ & - & 15.5 & 19.6 & 18.4 & 17.7 & 20.8 \\
        $Recall$    & - & 34.2 & 41.9 & 39.6 & 35.9 & 44.0 \\
        $F_1$        & 75.8 & 21.4 & 26.7 & 25.1 & 23.7 & 28.2 \\
    \hline
  \end{tabular}
  \caption{The 30-shot results of ChatGPT on the ACE05 benchmark compared to the supervised SOTA SaliencyED.}
  \label{tab:result-ee}
\end{table*}

\begin{table}[pt]
 \setcounter{table}{3}
  \centering
  \begin{tabular}{lcc}
    \toprule
        \textbf{Label} & \textbf{100} & \textbf{20000} \\
    \hline
        per:age & 0 & 1 \\
        per:cities\_of\_residence & 0 & 1 \\
        per:employee\_of & 1 & 1 \\
        no\_relation & 11 & 20 \\
        per:children & 0 & 1 \\
        org:top\_members/employees & 9 & 1 \\
        per:religion & 0 & 1 \\
        per:origin & 0 & 1 \\
        per:title & 9 & 2 \\
        org:country\_of\_headquarters & 0 & 1 \\
    \hline
  \end{tabular}
  \caption{The label distribution of demonstrations given by selection from candidate sets of size 100 and 20000.}
  \label{tab:result-re-case}
\end{table}

We randomly select 500 samples from the test set of TACRED to perform the validation. We split the results into 5 intervals according to the discrepancy scores and calculate the averages in the each interval. The results are shown in Figure~\ref{fig:example-re}.
We can observe that the RE performance of ChatGPT improves as the discrepancy values decreases, suggest the strong positive correlation between the content-based discrepancy and the performance.
It also demonstrates the generality of our proposed estimation method in IE tasks, as it only requires specifying an appropriate discrepancy metric.

\subsubsection{Experimental Results}


We evaluate the $30$-shot performance of our method with different size of demonstrations candidate set ranging from 100 to 20000.
The results are shown in table~\ref{tab:result-re}.
Note that it can be treated as randomly picking demonstrations when $n=100$.
Compare to random selection, demonstrations picked based on our method consistently boost the performance of ChatGPT by about 6\% in F1 score. As the size of demonstration candidate set goes large, the performance also becomes better. Even not directly comparable, our method roughly makes up a considerable proportion of the performance gap (78\% to 89\%) between chatGPT and SOTA methods.

\subsubsection{Case Study}

We compare the demonstrations selected from sets of size 100 (almost same as randomly selection) and 20000 (see Appendix~\ref{app:subsubsec:demos_re_carb}) as well as the result on the following sentence, where the entities are marked by "<entity> ... </entity>".
\begin{itemize}
    \item \textbf{Text}: \textit{Among those arrested was <entity> Wen Qiang </entity> , <entity> director </entity> of the municipal judicial administrative bureau and former executive deputy director of the municipal public security bureau.}
    \item \textbf{Label}: \textit{per:title}
\end{itemize}

In this case, we present the label frequencies of two sets of demonstrations in Table~\ref{tab:result-re-case}. Unlike the distribution of randomly sampled demonstrations biased towards the training set, the distribution resulting from our demonstrations more concentrate on the correct labels. This suggests that our method provides positive samples, thereby simplifying the corresponding task.

\subsection{Experiments on EE Task}

As an another ontology-specified IE task, Event Extraction (EE) aims at extracting the main content of structured event knowledge from given sentence. Based on the observation of the correlation validation in RE, we also specify the content-based discrepancy metric CaRB $F_1$ score in our method.

\subsubsection{Experimental Settings}

We conduct experiments on the event extraction benchmark ACE05~\citep{wadden2019entity}. It consists of 15,621, 893, and 727 samples for the train, development, and test sets, respectively. The objective is to classify each sample into one of 33 predefined event types. We randomly select 500 samples for evaluation and compare the results with the SOTA supervised method, SaliencyED~\citep{liu2022saliency}.
We use the same implementation to the RE experiment\footnote{Our experimental period is 2023.05.02-2023.06.08.}. The size of candidate set ranges from $50$ to $15000$.


\subsubsection{Experimental Results}


The results are shown in table~\ref{tab:result-ee}. Also, the setting of $n=50$ can be viewed as randomly demonstration selection. The experimental results show that our method enhance the performance of ChatGPT by a large margin (7.4\% in F1 score), which confirm the effectiveness of our method on EE task.
We also observe a certain disparity between the few-shot performance of our method and the supervised SOTA model. This finding suggests that ChatGPT may require further enhancement in its understanding capabilities for subjective tasks.


\section{Related Works}

\subsection{OIE Approaches}

OpenIE was first proposed by \citet{banko2007open} and has always been significant since it serves as the backbone of various downstream NLP tasks. A long line of work has attempted to build up practical OpenIE systems. The first kinds of approaches are statistic or rule-based models \citep{christensen2011analysis,schmitz2012open,del2013clausie,angeli2015leveraging, pal2016demonyms, saha2017bootstrapping}. These methods are generally based on heuristic predefined rules and statistic inference techniques. With the emergence of deep learning, a lot of supervised neural models have been proposed to handle OpenIE tasks. The first branches of works are tagging-based models \citep{stanovsky2018supervised, roy2019supervising, ro-etal-2020-multi, zhan2020span, kolluru2020openie6, yu2021maximal}. These approaches formulate OpenIE task as a sequence tagging task.
After the emergence of pretrained language models (PLM), several approaches also use PLMs to obtain context-aware embeddings. \citet{ro-etal-2020-multi} first used PLM and introduced a two-stage paradigm. \citet{kolluru2020openie6} further proposed an iterative grid labeling framework and used a transformer-based network to generate labels for each extraction in turn. Another branches of approaches are generative models \citep{cui2018neural, sun2018logician,kolluru2020imojie}, which formulate OpenIE task as a sequence-to-sequence generation task. \citet{cui2018neural} used three-layer LSTM for both encoder and decoder, and the encoder was later improved to BERT \citep{kolluru2020imojie}.

\subsection{IE based on LLMs}

The stunning chatting and instruction-following ability of ChatGPT naturally raises the question that to what extent can ChatGPT fulfill IE tasks. This question has elicited a line of concurrent works. Under zero-shot settings, \citet{li2023evaluating}, \citet{yuan2023zero} and \citet{wei2023zero} first explore the ability of ChatGPT on close IE tasks and found the performance to be disappointing. \citet{wei2023zero} further proposed a multi-stage framework to enhance the zero-shot ability. Under few-shot setting, \citet{han2023information} made a holistic evaluation of ChatGPT's performance on four IE tasks, while \citet{gao2023exploring} and \citet{xu2023unleash} focused on EE and RE, respectively. The results of these works reveal the significant performance gap between ChatGPT and SOTA. However, the demonstrations were selected randomly in all works listed above. \citet{wan2023gpt} first proposed a similarity-based demonstration selection method that enables ChatGPT to be comparable to SOTA but introduced an additional training cost. In this work, we focus on few-shot learning and propose demonstration selection methods only using off-the-shelf tools.



\subsection{Demonstrations Selection for LLMs}

The strong ability and hard-to-train nature of LLMs have led to a number of works studying demonstration selection for in-context learning based on similarity estimation. \citet{liu2022makes} proposed an in-context example retrieval algorithm to find the most semantically similar samples for NLU and generation tasks. Later supervising models \citep{das2021case, rubin2022learning} or reinforcement learning algorithms \citep{zhang2022active} were leveraged to retrieve demonstrations for various tasks. \citet{shin2021constrained} leveraged the LLM itself to rank the similarity.
For ChatGPT on IE tasks, \citet{wan2023gpt} makes the first approach to select similar demonstrations to enhance the performance but still requires further training.


\section{Conclusion}

In this paper, we study to enhance the few-shot reasoning capability of LLMs on specific tasks, by constructing a consistent reasoning environment to mitigate the distributional discrepancy between the test samples and LLMs. We first propose an estimation method to measure the distributional discrepancy between the black-box LLMs and test samples by employing a discrepancy metric, which can further derive the correlation between the discrepancy and the performance of the LLM on specific tasks. Guided by the estimation and the correlation, we further introduce a simple yet effective mechanism to prepare the reasoning environment with a majority of positive demonstrations may selected from a candidate set or crafted manually.
We conduct extensive experiments on the standard OIE benchmark CaRB. The results show that our $6$-shot approach outperforms the current state-of-the-art supervised model, achieving a $55.3$ $F_1$ score.  Further experiments on TACRED and ACE05 show that our method can naturally generalize to other information extraction tasks by replacing the discrepancy metric, resulting in improvements of $5.7$ Micro $F_1$ and $6.8$ $F_1$ scores, respectively.

\section{Limitations}

In this work, we propose a method for estimating the distribution discrepancy between test samples and large language models and preparing in-context learning environment to boost the performance of LLMs. However, it is not clear how to choose a appropriate distributional discrepancy metric at first glance. Meanwhile, our method though improves the performance of LLMs, still cannot fully make up the performance gap between LLMs and task-specific supervised neural models.

\section{Ethical and Broader Impacts}

In this paper, we propose a method to prepare reasoning environment with positive demonstrations and show that successfully selecting a set of consistent demonstrations will boost the performance of LLMs. However, manually designed biased or adversarial demonstrations might result in a performance degradation or a biased output. Also, our method encourages the usage of large scaled models, which might also aggravating the inequity between research groups with a plethora of computational resources or access to large language models and those lack of them.

\bibliography{emnlp2023, anthology}
\bibliographystyle{acl_natbib}

\clearpage
\appendix

\section{Appendix}
\label{sec:appendix}

\subsection{Prompts}
\label{app:prompts}


To facilitate reproducibility, we provide detailed prompts with accompanying examples. These prompts include the paraphrasing task, the sentence making task, the Open Information Extraction (OIE) task, the Relation Extraction (RE) task, and the Event Extraction (EE) task.

\subsubsection{Paraphrase Task}
\label{app:prompt_paraphrase}

For the paraphrasing task, we design the single-turn prompt in an instruction-style. We empirically find that utilizing this type of prompt can enable ChatGPT to generate paraphrase effectively.


\definecolor{'shallow1'}{HTML}{E4F3FC}
\begingroup
\begin{table}[ht]
    \centering
    \small
    \begin{tabular}{p{0.9\linewidth}}
        \toprule
            \vspace{-2mm}
        \cellcolor{'shallow1'} \textbf{\textsc{Instruction:}} \textit{Please give one paraphrase of the following sentence:  "They can be relieved only by changing that system , not by pouring Western money into it."  Please only output the sentence you made.} \\
        \midrule
        \vspace{-1mm}
        \cellcolor{'shallow1'} \textbf{\textsc{Output:}} \textit{The only way to alleviate their distress is by altering the system, rather than infusing it with Western funds.} \\
        \bottomrule
    \end{tabular}
        \caption{
  An example of promoting ChatGPT for the paraphrasing task, where "" enclose the input sentence.
    }
    \label{tab:instruction_1_1}
\end{table}
\endgroup

\subsubsection{Sentence Making Task}
\label{app:prompt_sent_making}


\definecolor{'shallow1'}{HTML}{E4F3FC}
\begingroup
\begin{table}[ht]
    \centering
    \small
    \begin{tabular}{p{0.9\linewidth}}
        \toprule
            \vspace{-2mm}
        \cellcolor{'shallow1'} \textbf{\textsc{Instruction:}} \textit{Given two phrases and a list of relations, please make several sentences such that in these sentences, the relation between the two phrases is close to one relation appears in the given list. If not possible or not enough context provided, please simply make sentences containing the two phrases. Make as many sentences as you can. Output the sentences only.} \\
        \midrule
        \vspace{-1mm}
        \cellcolor{'shallow1'} \textbf{\textsc{Demonstrations:}} \texttt{User}: \textit{Entities: Finn, Barry University. Relations: [org:founded, org:website, per:age, per:parents, per:employee\_of, no\_relation, org:parents, per:children, org:members, per:religion, per:origin, per:charges, per:title, org:dissolved, per:siblings, per:spouse, per:other\_family, per:city\_of\_death, org:founded\_by ...]}

\texttt{Assistant}: \textit{The Finn joined the coaching staff at Barry University in Miami Shores , Florida in 2001 , and then became an assistant at Florida State in 2005.
Finn is one of the most talented students in my college and thus has once been selected to be an exchanging student to Barry University.
Barry Unviersity is the first university I know, since once my neighbor Finn always says that his child was once admitted to it.
Finn is a high school student who has already received several offers from universities, among which the best offer is from Barry University.} \\
        \bottomrule
    \end{tabular}
        \caption{
  An example of promoting ChatGPT for the sentence making task.
    }
    \label{tab:instruction_1_1}
\end{table}
\endgroup

For the sentence making task, we design a prompt in a conversational style. We manually write a demonstration to guide the generation.

\subsubsection{OIE Task}
\label{app:prompt_oie}
For the OIE task, we design a prompt in a conversational style.

\definecolor{'shallow1'}{HTML}{E4F3FC}
\begingroup
\begin{table}[ht]
    \centering
    \small
    \begin{tabular}{p{0.9\linewidth}}
        \toprule
            \vspace{-2mm}
        \cellcolor{'shallow1'} \textbf{\textsc{Instruction:}} \textit{Open information extraction requires the extraction of all relations in the sentence, i.e., predicates, the subjects and objects corresponding to these relations, and the possible time and place. The results should be display in the format of tuples. In these tuples, we always put the predicate first, the second is the subject corresponding to the predicate, the third is the object corresponding to the predicate (if there is none, it is not labeled), and the last two are time and place, which should be omitted if there is none. Please extract information tuples from the following sentences and show the results in one line.} \\
        \midrule
        \vspace{-1mm}
        \cellcolor{'shallow1'} \textbf{\textsc{Demonstrations:}} \texttt{User}: \textit{[demonstrations]>}

\texttt{Assistant}: \textit{[demonstrations]} \\
        \bottomrule
    \end{tabular}
        \caption{
  An example of promoting ChatGPT for the sentence making task.
    }
    \label{tab:instruction_1_1}
\end{table}
\endgroup


\subsubsection{RE Task}
\label{app:prompt_re}

For the RE task, we design a prompt in a conversational style.

\definecolor{'shallow1'}{HTML}{E4F3FC}
\begingroup
\begin{table}[ht]
    \centering
    \small
    \begin{tabular}{p{0.9\linewidth}}
        \toprule
            \vspace{-2mm}
        \cellcolor{'shallow1'} \textbf{\textsc{Instruction:}} \textit{Please classify relationships between the two entities (marked with <entity> and </entity>). The set of relationships is as follows: [org:founded, org:subsidiaries, per:date\_of\_birth, per:cause\_of\_death, per:age, per:stateorprovince\_of\_birth, per:countries\_of\_residence, per:country\_of\_birth, per:stateorprovinces\_of\_residence, org:website, per:cities\_of\_residence, per:parents, per:employee\_of, no\_relation, per:city\_of\_birth, org:parents, org:political/religious\_affiliation, per:schools\_attended, per:country\_of\_death, per:children, org:top\_members/employees, per:date\_of\_death, org:members, org:alternate\_names, per:religion, org:member\_of, org:city\_of\_headquarters, per:origin, org:shareholders, per:charges, per:title, org:number\_of\_employees/members, org:dissolved, org:country\_of\_headquarters, per:alternate\_names, per:siblings, org:stateorprovince\_of\_headquarters, per:spouse, per:other\_family, per:city\_of\_death, per:stateorprovince\_of\_death, org:founded\_by].} \\
        \midrule
        \vspace{-1mm}
        \cellcolor{'shallow1'} \textbf{\textsc{Demonstrations:}} \texttt{User}: \textit{[demonstrations]}

\texttt{Assistant}: \textit{[demonstrations]} \\
        \bottomrule
    \end{tabular}
        \caption{
  An example of promoting ChatGPT for the sentence making task.
    }
    \label{tab:instruction_1_1}
\end{table}
\endgroup


\subsubsection{EE Task}
\label{app:prompt_ee}

For the EE task, we design a prompt in a conversational style.

\definecolor{'shallow1'}{HTML}{E4F3FC}
\begingroup
\begin{table}[ht]
    \centering
    \small
    \begin{tabular}{p{0.9\linewidth}}
        \toprule
            \vspace{-2mm}
        \cellcolor{'shallow1'} \textbf{\textsc{Instruction:}} \textit{Please identify the words that indicating events in the text and classify them into appropriate categories; The collection of categories is [Conflict.Attack, Movement.Transport, Life.Die, Contact.Phone-Write, Life.Injure, Contact.Meet, Transaction.Transfer-Ownership, Personnel.End-Position, Justice.Arrest-Jail, Conflict.Demonstrate, Life.Marry, Personnel.Elect, Personnel.Start-Position, Personnel.Nominate, Business.End-Org, Justice.Execute, Business.Start-Org, Justice.Fine, Transaction.Transfer-Money, Justice.Trial-Hearing, Justice.Sue, Justice.Charge-Indict, Justice.Sentence, Life.Be-Born, Justice.Extradite, Business.Declare-Bankruptcy, Justice.Convict, Justice.Release-Parole, Business.Merge-Org, Justice.Appeal, Justice.Pardon, Life.Divorce, Justice.Acquit]} \\
        \midrule
        \vspace{-1mm}
        \cellcolor{'shallow1'} \textbf{\textsc{Demonstrations:}} \texttt{User}: \textit{[demonstrations]}

\texttt{Assistant}: \textit{[demonstrations]} \\
        \bottomrule
    \end{tabular}
        \caption{
  An example of promoting ChatGPT for the sentence making task.
    }
    \label{tab:instruction_1_1}
\end{table}
\endgroup


\subsection{Detailed Case Study}\label{app:case}

In this section we present the detailed demonstrations for case study in OpenIE and RE.
For OpenIE, the randomly selected demonstrations and the demonstrations selected according to syntactic distance are shown in Sec.~\ref{app:subsubsec:demos_oie_random} and Sec.~\ref{app:subsubsec:demos_oie_hws} respectively.
For RE task, the randomly picked demonstrations and the demonstrations selected according to CaRB metric are shown in Sec.~\ref{app:subsubsec:demos_re_random} and Sec.~\ref{app:subsubsec:demos_re_carb}:

\subsubsection{Demonstrations Selected Randomly}
\label{app:subsubsec:demos_oie_random}

\begin{itemize}
    \item \texttt{SENTENCE}: Leaving only a small contigent to guard the defile, he took the entire army to destroy the plain lying ahead of Alexander's army. \texttt{EXTRACTION}: (took, he, the entire army);(was to destroy, the entire army, the plain lying ahead of Alexander's army);(was lying ahead of, the plain, Alexander's army);(was to guard, a small contigent, the defile)
    \item \texttt{SENTENCE}: The bonds of New York, after being hammered for weeks due to the pending supply and reports that the city's economy is weakening, rose 1/2 point yesterday. \texttt{EXTRACTION}: (was hammered, The bonds of New York, for weeks due to the pending supply);(reports, The bonds of New York, that the city's economy is weakening, rose 1/2 point yesterday)
    \item \texttt{SENTENCE}: On a recent afternoon, Mr. Baker and a reporter go ghost-busting, visiting Kathleen Stinnett, a woman from Lexington who called the University of Kentucky to report on mysterious events at her house. \texttt{EXTRACTION}: (go ghost-busting, Mr. Baker and a reporter, on a recent afternoon);(visit, Mr. Baker and a reporter, Kathleen Stinnett, on a recent afternoon);(is, Kathleen Stinnett, a woman from Lexington);(called, Kathleen Stinnett, the University of Kentucky to report on mysterious events, at her house.)
\end{itemize}

\subsubsection{Demonstrations Selected Based HWS metric}
\label{app:subsubsec:demos_oie_hws}

\begin{itemize}
    \item \texttt{SENTENCE}: Software written for other Mips computers can also run on this machine, the company said. \texttt{EXTRACTION}: (written for other Mips computers, Software, can also run, on this machine,, the company, said.);(can also run, Software, on this machine,, the company, said.);(is written, Software, for other Mips computers);(the company, said)
    \item \texttt{SENTENCE}: The fitness craze itself has gone soft , the survey found.  \texttt{EXTRACTION}: (has gone soft, The fitness craze has);(found, the survey, The fitness craze itself has gone soft);(is of fitness, The craze)
    \item \texttt{SENTENCE}: Large cross - border deals numbered 51 and totaled \$ 17.1 billion in the second quarter , the firm added. \texttt{EXTRACTION}: (numbered, Large cross - border deals, 51, in the second quarter);(totaled, Large cross - border deals, \$ 17.1 billion, in the second quarter);(added, the firm, Large cross - border deals numbered 51 and totaled \$ 17.1 billion, in the second quarter)
\end{itemize}

\subsubsection{Demonstrations Selected Randomly}
\label{app:subsubsec:demos_re_random}

\begin{itemize}
    \item \texttt{SENTENCE}: Teenage defendant <entity> Brandon McInerney </entity> of <entity> Oxnard </entity> is charged with first-degree murder and a hate crime in connection with the Feb 12 killing of classmate Larry King , 15 , who sometimes wore makeup and told friends he was gay . \texttt{LABEL}: per:cities\_of\_residence
    \item \texttt{SENTENCE}: Claiming tens of millions of members , the National PTA , <entity> National Education Association </entity> , Parents Choice Foundation , the YWCA USA , the <entity> National Military Family Association </entity> and other advocacy groups joined the council in announcing the formation of the `` Smart Television Alliance . ''
     \texttt{LABEL}: no\_relation
    \item \texttt{SENTENCE}: <entity> He </entity> added that some <entity> 1,000 </entity> jobs have been trimmed off the payroll either through layoffs or attrition , and that the airline will continue to freeze hiring and cut staff through natural attrition .
     \texttt{LABEL}: no\_relation
    \item \texttt{SENTENCE}: Writers like <entity> William Anderson </entity> and <entity> Dorothy Rabinowitz </entity> have pointed out the parallel between the Duke case and earlier child abuse hysteria , where feminist prosecutors whipped up public invective against parents they knew to be innocent .
     \texttt{LABEL}: no\_relation
    \item \texttt{SENTENCE}: Persad-Bissessar , from the <entity> UNC </entity> which largely relies on <entity> Indo-Trinidadian </entity> backing , is seeking multi-ethnic support in her `` People 's Partnership '' .
     \texttt{LABEL}: no\_relation
    \item \texttt{SENTENCE}: <entity> He </entity> and <entity> his </entity> group also joined in a legal battle challenging the Washington Redskins ' trademarked name .
     \texttt{LABEL}: no\_relation
    \item \texttt{SENTENCE}: Miranda Lambert made history Wednesday morning when she was nominated for <entity> nine </entity> <entity> CMA </entity> Awards , the most for a female country music artist .
     \texttt{LABEL}: no\_relation
    \item \texttt{SENTENCE}: This surge in the number of cases has overwhelmed us , '' <entity> Sutedja </entity> was quoted by the <entity> Jakarta Globe </entity> as saying .
     \texttt{LABEL}: no\_relation
    \item \texttt{SENTENCE}: <entity> He </entity> rose to <entity> captain </entity> and served in the Pacific theater on the staff of Gen Douglas MacArthur .
     \texttt{LABEL}: per:title
    \item \texttt{SENTENCE}: In spite of a colourful family background , <entity> Alice </entity> , <entity> 20 </entity> , has so far managed to keep her reputation intact - but she now seems to have shed her clean-living image .
     \texttt{LABEL}: per:age
    \item \texttt{SENTENCE}: GREECE \_ BRUSSELS <entity> \_ </entity> European chiefs are putting the <entity> International Monetary Fund </entity> on standby to aid debt-stricken Greece , seeking to snuff out a threat to the stability of the euro .
     \texttt{LABEL}: no\_relation
    \item \texttt{SENTENCE}: Taiwan 's <entity> Defence </entity> Minister <entity> Lee Jye </entity> said last week the island needed to buy more advanced weaponry to counter the threat from China 's rapid military buildup , which he said had seen Taiwan gradually lose its air and naval superiority .
     \texttt{LABEL}: no\_relation
    \item \texttt{SENTENCE}: <entity> President </entity> Jalal Talabani , a Kurd , said former Defense Minister Sultan Hashim Ahmad al-Tai deserved to be spared because <entity> he </entity> had been carrying out orders under threat of death by Saddam and because he had engaged in official contact with the Kurdish community under the ousted regime .
     \texttt{LABEL}: no\_relation
    \item \texttt{SENTENCE}: It also is examining the disputed kidnapping of Khalid el-Masri , a German citizen of Lebanese descent , and the detention of <entity> Murat Kurnaz </entity> , a German - born <entity> Turkish </entity> national .
     \texttt{LABEL}: per:origin
    \item \texttt{SENTENCE}: Citing privacy laws , U.S. Citizenship and Immigration Services <entity> spokesman </entity> <entity> Chris Bentley </entity> declined to comment specifically on the Campbells ' case .
     \texttt{LABEL}: per:title
    \item \texttt{SENTENCE}: Japan 's Pentax Corp. may seek monetary damages from a battery - making subsidiary of electronics giant Matsushita after a fire at a plant run by the company caused <entity> Pentax </entity> to delay the launch of a new camera , Pentax said <entity> Monday </entity> , according to a media report .
     \texttt{LABEL}: no\_relation
    \item \texttt{SENTENCE}: Prosecutors say that <entity> Brandon McInerney </entity> killed gay teen Lawrence King in part because of the influence of neo-Nazis who had befriended <entity> him </entity> .
     \texttt{LABEL}: no\_relation
    \item \texttt{SENTENCE}: Tehran , Iran , Aug. 02 -- Iran 's Supreme Leader <entity> Ayatollah Ali Khamenei </entity> blasted United States policy in the Middle East and warned of an impending <entity> Muslim </entity> `` jihad '' , or holy war , against the West .
     \texttt{LABEL}: per:religion
    \item \texttt{SENTENCE}: <entity> Mitchell </entity> was elected to the <entity> U.S. House of Representatives </entity> from Baltimore in 1970 and was Maryland 's first African-American congressman .
     \texttt{LABEL}: per:employee\_of
    \item \texttt{SENTENCE}: During his first tour as chief executive , <entity> Brown </entity> led <entity> MBIA </entity> into the risky business of insuring complex mortgage-related investments -- the very investments now threatening to cripple the company .
     \texttt{LABEL}: org:top\_members/employees
    \item \texttt{SENTENCE}: `` It appears that deterrence has been restored , '' said <entity> Daniel Pinkston </entity> , Seoul-based analyst with the <entity> International Crisis Group </entity> think tank .
     \texttt{LABEL}: no\_relation
    \item \texttt{SENTENCE}: The <entity> ICG </entity> said accusations that <entity> Khartoum </entity> is instigating tribal clashes in the south are `` unsubstantiated '' , urging the government of south Sudan to `` focus internally . ''
     \texttt{LABEL}: no\_relation
    \item \texttt{SENTENCE}: Juventud <entity> 0 </entity> , <entity> Defensor </entity> 2
     \texttt{LABEL}: no\_relation
    \item \texttt{SENTENCE}: Mary Lynne Bird , the current executive director of the society , said <entity> Nolte </entity> led negotiations with the University of Wisconsin in <entity> 1978 </entity> when ownership of the society 's collection of maps , journals by explorers , artifacts from explorations and surveys went to the university .
     \texttt{LABEL}: no\_relation
    \item \texttt{SENTENCE}: Of course , <entity> Ezra Levant </entity> had to dash down to Waterloo and start poking around , amazed that the Waterloo police chief would n't release further details to him and <entity> his </entity> adoring fans .
     \texttt{LABEL}: no\_relation
    \item \texttt{SENTENCE}: <entity> Independent Steelworkers Union </entity> formally merges with USW One of the last independent steel unions in the <entity> US </entity> formally merged with the United Steelworkers on Friday , adding to the ranks of North America 's largest industrial labor union .
     \texttt{LABEL}: org:country\_of\_headquarters
    \item \texttt{SENTENCE}: `` We expect them not to make a hasty judgment , '' said <entity> Haddadadel </entity> , who leads the Abadgaran Iran-e-Islami party , or <entity> Developers of Islamic Iran </entity> , which is expected to set the tone in the new parliament that convenes in June .
     \texttt{LABEL}: no\_relation
    \item \texttt{SENTENCE}: <entity> Her </entity> daughter will live with the singer , the <entity> Evening Standard </entity> newspaper reported .
     \texttt{LABEL}: no\_relation
    \item \texttt{SENTENCE}: <entity> Kollek </entity> is survived by his widow Tamar , son Amos and daughter <entity> Osnat </entity> .
     \texttt{LABEL}: per:children
    \item \texttt{SENTENCE}: <entity> Victorino </entity> 's father , Mike , who flew in from Maui for the division series , told the <entity> Philadelphia Inquirer </entity> last week that this was the perfect city for his son because it is a blue-collar town and Shane is a blue-collar guy .
     \texttt{LABEL}: no\_relation
\end{itemize}

\subsubsection{Demonstrations Selected based on CaRB metric}
\label{app:subsubsec:demos_re_carb}

\begin{itemize}
    \item \texttt{SENTENCE}: `` Raising the minimum wage will cost some jobs , '' <entity> Al Hubbard </entity> , the <entity> director </entity> of the president 's National Economic Council told reporters aboard Air Force One Tuesday . Label: no\_relation
    \item \texttt{SENTENCE}: `` It started so early this year that we have pictures of some of the candidates wearing shorts , '' said <entity> Paul Manuel </entity> , executive director of the <entity> New Hampshire Institute of Politics </entity> at St. Anselm College in Manchester .
     \texttt{LABEL}: no\_relation
    \item \texttt{SENTENCE}: Among those who received a 15-year sentence was former Defense <entity> Minister Sultan </entity> <entity> Hashim Ahmad al-Tai </entity> , who signed the cease-fire with U.S.-led forces that ended the 1991 war .
     \texttt{LABEL}: per:title
    \item \texttt{SENTENCE}: `` Our regulations will help prevent unsafe activity within the cab , '' <entity> Anne Ferro </entity> , head of the <entity> Federal Motor Carrier Safety Administration </entity> , said in the statement .
     \texttt{LABEL}: org:top\_members/employees
    \item \texttt{SENTENCE}: It 's helpful to be prepared to talk about the mythology and history of Santa Claus , said <entity> Frank Farley </entity> , former president of the <entity> American Psychological Association </entity> .
     \texttt{LABEL}: org:top\_members/employees
    \item \texttt{SENTENCE}: `` By and large , American schools and campuses are safe places , '' said <entity> Irwin Redlener </entity> , director of the <entity> National Center for Disaster Preparedness </entity> , at Columbia University .
     \texttt{LABEL}: org:top\_members/employees
    \item \texttt{SENTENCE}: `` Any team in smaller markets should be encouraged to expand their market , '' said <entity> Robert McNair </entity> , the chief <entity> executive </entity> of the Houston Texans .
     \texttt{LABEL}: per:title
    \item \texttt{SENTENCE}: `` It 's another attempt to prop up a black face and say , ` See , we 're OK , '' ' said <entity> Julian Bond </entity> , chairman of the <entity> NAACP </entity> .
     \texttt{LABEL}: per:employee\_of
    \item \texttt{SENTENCE}: Among the letter writers was Terry Stevens of <entity> Francesville </entity> , Ind. , who identified himself as a bank investor , as the <entity> ABA </entity> had suggested .
     \texttt{LABEL}: no\_relation
    \item \texttt{SENTENCE}: Among those who believe the commotion over the drug is overblown is <entity> Rick Doblin </entity> of the <entity> Multidisciplinary Association for Psychedelic Studies </entity> , a nonprofit group that does research on psychedelic drugs and whose goal is to develop psychedelics and marijuana into prescription medication .
     \texttt{LABEL}: no\_relation
    \item \texttt{SENTENCE}: `` We hope to have the roads back to normal in a few days , '' said <entity> Rick Sullivan </entity> , commissioner of the <entity> Department of Conservation and Recreation </entity> .
     \texttt{LABEL}: no\_relation
    \item \texttt{SENTENCE}: Of course we welcome this timeframe , '' said <entity> Abdul Jalil </entity> , general secretary of the <entity> Awami League </entity> .
     \texttt{LABEL}: org:top\_members/employees
    \item \texttt{SENTENCE}: `` We realize employers are n't forensic document examiners , '' said <entity> Bentley </entity> , the USCIS <entity> spokesman </entity> .
     \texttt{LABEL}: per:title
    \item \texttt{SENTENCE}: `` We 're not taking any chances this time , '' said <entity> Sam Williams </entity> , president of the <entity> chamber </entity> .
     \texttt{LABEL}: org:top\_members/employees
    \item \texttt{SENTENCE}: `` Now we know why it seemed like Mexican officials knew where we were all the time , '' said <entity> Chris Simcox </entity> , founder of the <entity> Minuteman Civil Defense Corps </entity> .
     \texttt{LABEL}: no\_relation
    \item \texttt{SENTENCE}: They are important in that they provide affordable housing and they keep these communities diverse , '' said <entity> Larry Gross </entity> , executive director of the <entity> Coalition for Economic Survival </entity> .
     \texttt{LABEL}: org:top\_members/employees
    \item \texttt{SENTENCE}: Democrat <entity> Chris Dodd </entity> , <entity> chairman </entity> of the Senate banking committee , lamented the breakdown , particularly with the holidays approaching .
     \texttt{LABEL}: per:title
    \item \texttt{SENTENCE}: <entity> Christopher Bentley </entity> , <entity> spokesman </entity> for the US
     \texttt{LABEL}: per:title
    \item \texttt{SENTENCE}: Among those who received a 15-year sentence was former Defense Minister Sultan <entity> Hashim Ahmad al-Tai </entity> , who signed the cease-fire with <entity> U.S.-led </entity> forces that ended the 1991 war .
     \texttt{LABEL}: no\_relation
    \item \texttt{SENTENCE}: `` He spent a lot of time listening , but he opened it by saying that they were going to push hard to really get this bill done and get it done in the near term , '' said <entity> Edward L. Yingling </entity> , president of the <entity> American Bankers Association </entity> .
     \texttt{LABEL}: org:top\_members/employees
    \item \texttt{SENTENCE}: Among the alumni is Connecticut Sen. <entity> Christopher J. Dodd </entity> , who served as a volunteer in the <entity> Dominican Republic </entity> .
     \texttt{LABEL}: no\_relation
    \item \texttt{SENTENCE}: Roundup : Iranian conservative camp preaches `` justice '' for presidential elections The first possible nominee is <entity> Ali Larijani </entity> , former <entity> director </entity> of the Islamic Republic of Iran Broadcasting -LRB- IRIB -RRB- .
     \texttt{LABEL}: per:title
    \item \texttt{SENTENCE}: Among the letter writers was <entity> Terry Stevens </entity> of Francesville , Ind. , who identified himself as a bank investor , as the <entity> ABA </entity> had suggested .
     \texttt{LABEL}: no\_relation
    \item \texttt{SENTENCE}: <entity> Abdul Karim al-Khawinay </entity> , <entity> editor </entity> of the Al-Shura weekly , was abducted on the street by about 7 people who jumped out of a land cruiser with covered car plates and took him away , the Yemeni Journalists Syndicate said , citing eyewitnesses .
     \texttt{LABEL}: per:title
    \item \texttt{SENTENCE}: Among other auctioned items was a photo of Sills as Queen <entity> Elizabeth </entity> that graced a 1971 cover of Time magazine , which called <entity> her </entity> `` America 's queen of opera . ''
     \texttt{LABEL}: no\_relation
    \item \texttt{SENTENCE}: Among those in attendance was 27-year-old <entity> Navy Lieutenant </entity> <entity> John Kerry </entity> , who had served on a Swift Boat in Vietnam .
     \texttt{LABEL}: per:title
    \item \texttt{SENTENCE}: <entity> Sharpton </entity> is president of the <entity> National Action Network </entity> .
     \texttt{LABEL}: org:top\_members/employees
    \item \texttt{SENTENCE}: Among the notable chin-scratchers was the case against an <entity> Ontario </entity> surgeon who refused to perform cosmetic plastic surgery on the vagina of a trangendered woman because he had no experience with sex-changes , and the Sikh man who was supported by the <entity> Ontario Human Rights Commission </entity> in his bid to exempt himself from the law that motorcyclists must wear helmets , because it would mean removing his turban .
     \texttt{LABEL}: no\_relation
    \item \texttt{SENTENCE}: `` None of us were embarrassed '' by the effort to pursue a broader health agenda , said <entity> Al Hubbard </entity> , <entity> director </entity> of the National Economic Council at the White House .
     \texttt{LABEL}: per:title
    \item \texttt{SENTENCE}: `` We now see this as a way to help the package pass , '' said <entity> Ed Yingling </entity> , president of the <entity> American Bankers Association </entity> .
     \texttt{LABEL}: org:top\_members/employees
\end{itemize}

\end{document}